\documentclass[runningheads]{llncs}

\usepackage{eccv}

\usepackage{eccvabbrv}

\usepackage{graphicx}
\usepackage{booktabs}
\usepackage{fontawesome5}

\usepackage[accsupp]{axessibility}  % Improves PDF readability for those with disabilities.

\usepackage[pagebackref,breaklinks,colorlinks,citecolor=eccvblue]{hyperref}
\usepackage{orcidlink}

\newcommand{\ours}{Curia-MAE}

\begin{document}

% ---------------------------------------------------------------
\title{\ours{}: Multi-Modal Multi-Anatomy MAE Pre-Training for 3D Medical Image Segmentation} 

\titlerunning{Multi-Modal Multi-Anatomy MAE Training for Medical Segmentation}

\author{Théo Danielou\thanks{Equal contribution.}\inst{1}\orcidlink{0009-0001-4414-4781} \and
Antoine Saporta\protect\footnotemark[1]\inst{1}\orcidlink{0009-0002-2308-6962} \and Léo Alberge\inst{1}\orcidlink{0009-0002-6633-3791} \and Corentin Dancette\inst{1}\orcidlink{0009-0007-9274-4910}}

\authorrunning{T.~Danielou et al.}

\institute{Raidium \\
\email{firstname.lastname@raidium.eu}}

\maketitle

\begin{abstract}
Radiology foundation models learn transferable representations that can be adapted to new tasks by training only small layers on top of a frozen encoder. Dense prediction tasks such as 3D segmentation are, however, underrepresented in their evaluation, and, with the encoder kept frozen, pre-trained models still fall short of nnU-Net, the state-of-the-art reference trained from scratch. To close this gap we extend convolutional MAE pre-training with a robust reconstruction objective, a feature regularizer, and a local--global similarity objective. Using this method, we propose \ours{}, a multi-modal, multi-anatomy MAE model pre-trained on 300,000 CT and MRI images covering a large number of anatomical sites.
%We propose a Multi-Modal Multi-Anatomy Masked AutoEncoder (\ours{}), pre-trained on 300,000 CT and MRI images covering a large number of anatomical sites. \ours{} extends a convolutional MAE with a robust reconstruction objective, a feature regularizer, and a local--global similarity objective. 
On eight anatomy- and lesion-focused segmentation benchmarks, \ours{} improves frozen-encoder performance over a strong MAE baseline, while remaining competitive under full finetuning and superior on lesion tasks, where labeled data is scarce. These results indicate that a single frozen encoder can be reused across diverse segmentation tasks, reducing the cost of adapting and deploying such models in clinical workflows. Curia-MAE pre-trained model weights are made publicly available at \url{https://huggingface.co/raidium/Curia-MAE}.

\end{abstract}

\section{Introduction}
\label{sec:intro}
Medical imaging is used across clinical specialties, from initial diagnosis through treatment monitoring. Historically, applying AI to support radiologist analysis required a patchwork of specialized models, each solving a single, specific task. Foundation Models (FMs) reframe the problem: by distilling broad, transferable representations from large image collections, a single pre-trained encoder can be repurposed across a variety of downstream applications at minimal cost, by training only small linear layers on top of the frozen encoder. Recent research in the medical imaging community has delivered a broad spectrum of medical foundation models~\cite{codella2024MedImageInsight, zhang2025BioMedCLIP, sellergren2025medgemma, blankemeier2024merlin, agrawal2025pillar, dancette2025curia, saporta2026curia}, shown to address a wide variety of practical clinical tasks across modalities and anatomical sites.  

Yet, despite their central role in radiological workflows, segmentation tasks are largely absent from the evaluation benchmarks of those foundation models. Worse, recent works~\cite{munk2024amaes,revisitingmae} suggest that models pre-trained for such dense tasks still fall short of nnU-Net~\cite{nnunet}, the current state-of-the-art reference in medical segmentation, when only a dedicated decoder is trained on top of the frozen encoder. Conceptually, a strong backbone usable in a frozen encoder setting would be valuable: adapting to a new task by training only a decoder would require less compute and labeled data, cutting costs. It would also reshape deployment of segmentation models in real clinical applications, as sharing a single encoder across the many models a radiologist runs in its routine workflows would amortize both inference time and compute cost.

In this work, we propose a follow up on the research initiated by \cite{revisitingmae} on Masked Auto-Encoders (MAE) and tackle two key limitations of that work: (1) the frozen-encoder setting is reported to be detrimental to segmentation training; and (2) their pre-training is restricted to a single modality and anatomical region, namely head-and-neck MRI. Our contributions are three-fold:
\begin{enumerate}
    \item Building on the strategy of~\cite{revisitingmae}, we propose a new pre-training methodology that improves the capabilities of the pre-trained encoder, in particular in a frozen-encoder setting. More specifically, the proposed improvements include changes to the MAE reconstruction loss, as well as the introduction of feature regularization and a local--global similarity objective. We ablate each of the proposed methodological contributions.
    \item Using this methodology, we train our model on a radiological dataset of roughly 300,000 volumes, spanning both CT and MRI and covering a range of anatomical regions. We name this model \ours{}, a multi-modal, multi-anatomy MAE model.
    \item We evaluate both \ours{} and a MAE baseline on a suite of downstream segmentation tasks, covering both anatomy and lesion-focused targets across modalities, and compare against the nnU-Net state of the art. We further extend the evaluation to classification and regression tasks using the CuriaBench3D~\cite{saporta2026curia} benchmark.
\end{enumerate}

\section{Related Works}
Within the broader computer vision landscape, self-supervised learning (SSL) encompasses, on one hand, frameworks focused on learning robust visual features through instance-level contrastive learning or self-distillation paradigms, such as DINO \cite{dino}, BYOL \cite{byol}, Barlow Twins \cite{barlowtwins}, and CLIP \cite{clip}. Recently, this paradigm has been successfully adapted to the clinical domain to build large-scale radiology foundation models. Prominent examples include architectures like BioMedCLIP \cite{zhang2025BioMedCLIP} and the Curia \cite{dancette2025curia} framework, alongside its recent evolution Curia-2 \cite{saporta2026curia}, which scales representation quality to billion-parameter models using massive multimodal CT and MRI datasets. Nevertheless, the optimization objectives of these foundation models heavily prioritize global or slice-level features, image classification, and multi-finding screening. Lacking an inherent mechanism for dense, voxel-wise feature localization, such architectures yield representations that are poorly suited for dense downstream prediction tasks, most notably volumetric semantic segmentation.

Masked Autoencoders (MAE) \cite{mae} have emerged as a highly scalable and effective self-supervised representation learning paradigm for dense downstream tasks. While initially tailored for Transformer architectures due to their inherent sequence modeling capabilities, subsequent advancements successfully adapted MAE to Convolutional Neural Networks (CNNs) by modifying standard convolutional operations and normalization schemes to remain resilient to spatial masking \cite{tian2023designing} \cite{woo2023convnext}. In the medical imaging domain, initial applications of the MAE framework heavily favored 3D Vision Transformers employing direct masking and reconstruction objectives \cite{tang2022self} \cite{chen2023masked} \cite{zhuang2025advancing}. However, CNNs remain the cornerstone of medical image processing; most notably, frameworks like nnU-Net~\cite{nnunet} have robustly demonstrated that convolutional architectures consistently deliver state-of-the-art performance across a wide range of medical segmentation benchmarks. Driven by this proven efficacy, the field has recently transitioned toward bringing the MAE paradigm to 3D medical CNNs, with contemporary work demonstrating the success of volumetric masking and reconstruction directly on convolutional backbones \cite{munk2024amaes}.

Building upon this momentum, recent efforts have sought to formalize and optimize volumetric MAE pre-training. As highlighted in \cite{revisitingmae}, prior 3D medical SSL literature has historically suffered from three primary limitations: constrained dataset sizes, a reliance on transformer backbones that lag behind well-configured CNNs, and inadequate validation practices. To address these pitfalls, they leverage roughly 39,000 3D head-and-neck MRI volumes to scale the pre-training regime, introducing an optimized 3D CNN MAE paradigm that sets a new state-of-the-art. However, certain limitations persist. The approach remains restricted to a single anatomical region and modality. Furthermore, their scaling ablations reveal an unresolved bottleneck, as extending training duration or increasing batch size failed to yield downstream performance gains. Finally, consistent with conventional MAE approaches, keeping the encoder weights frozen during evaluation proved highly detrimental, highlighting that full fine-tuning remains essential for effective representation transfer.

\section{Method}
\label{sec:method}
\begin{figure}[t]
    \centering
    \includegraphics[width=\linewidth]{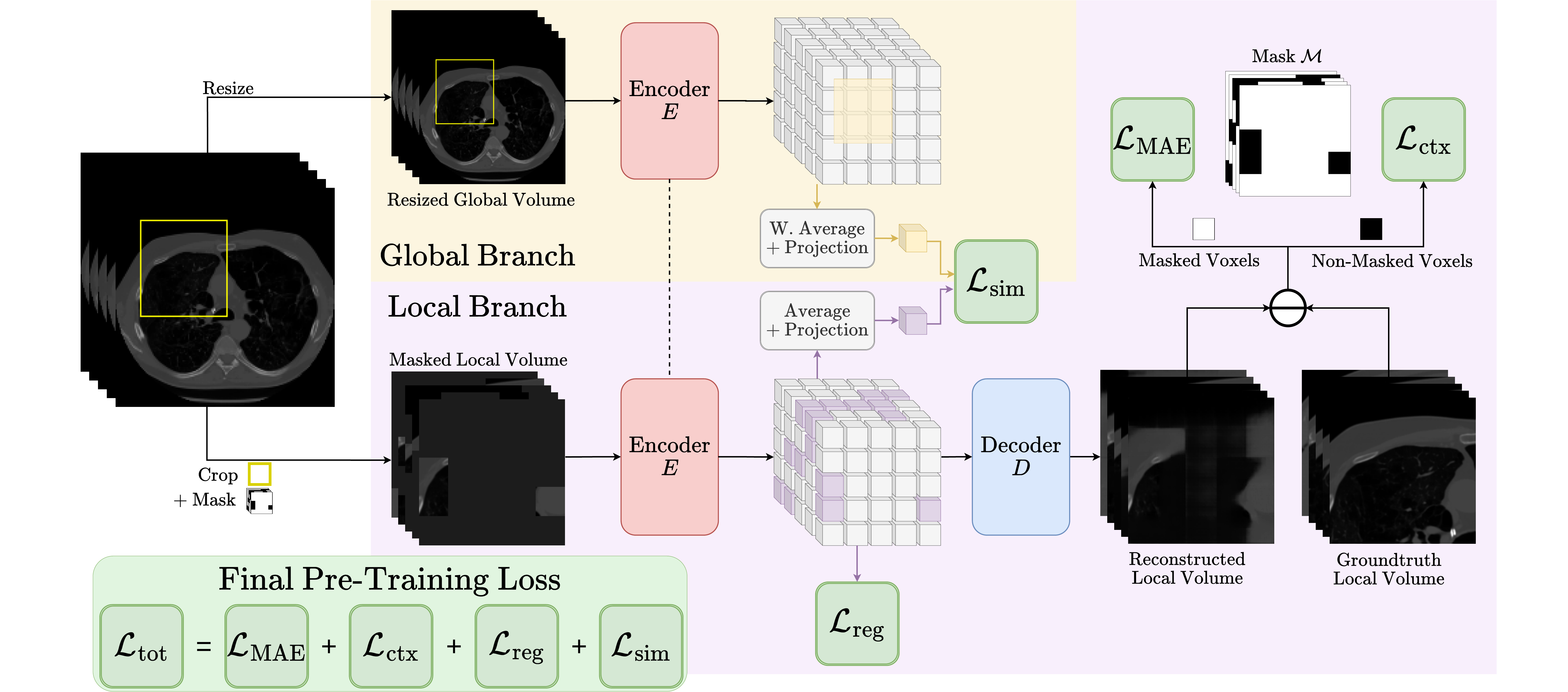}
    \caption{\textbf{Overview of our method.} A 3D volume is fed into two branches. On the one hand, a cropped, masked view of the volume is processed in a local branch to perform masked auto-encoding: the encoder-decoder learns to reconstruct the groundtruth local volume by inferring both masked voxels ($\mathcal{L}_\text{MAE}$) and non-masked voxels ($\mathcal{L}_\text{ctx}$). Features at the bottleneck are further regularized to prevent collapse to a low-dimensional subspace ($\mathcal{L}_\text{reg}$). On the other hand, the full 3D volume, reshaped to the encoder's expected input shape, is forwarded to extract the features matching the local crop. A local-global similarity objective ensures that semantically equivalent inputs yield consistent representations ($\mathcal{L}_\text{sim}$).}
    \label{fig:method}
\end{figure}

This section details the methodological improvements we add to the MAE strategy for medical imaging proposed in \cite{revisitingmae}. Our method, leading to \ours{}, is illustrated in Fig.~\ref{fig:method}. Our contributions aim to tackle a couple of limitations of \cite{revisitingmae}: (1) the original work was limited to head-and-neck MRI; and (2) it reported that keeping the encoder weights frozen in downstream training is detrimental, whereas a strong frozen-encoder representation is a desirable property of SSL-pretrained models. In Section~\ref{sec:experiments}, we describe how we pre-train models on a multimodal, multi-anatomical dataset and evaluate them on publicly available benchmarks; each component of our method is ablated in Section~\ref{sec:ablation}.

\subsubsection{Masked Auto-Encoder.}
The Masked Autoencoder (MAE) self-supervised learning paradigm involves masking a specific proportion of the input image patches and training the model to reconstruct the corrupted regions. Although MAE was originally conceptualized for Vision Transformer (ViT) architectures, we adapted this framework for a Residual Encoder (ResEnc) backbone~\cite{isensee2024nnu}, following the methodology proposed by \cite{revisitingmae}. This adaptation incorporates sparse convolutions, specialized sparse normalizations, mask tokens, and a subsequent convolutional densification stage to accommodate convolutional networks. The architectural skip connections are maintained during the pre-training phase, following \cite{revisitingmae}.
\paragraph{Baseline.} As initial baseline, we apply the MAE framework of \cite{revisitingmae} on our pre-training dataset. More specifically, we first apply a similar pre-processing to the volumetric data, including using $z$-score normalization, resampling images to a target spacing of $(1.0,1.0,1.0)$, and an input patch size of $P^3=160\times 160\times 160$. The mask $\mathcal{M}\in\{0,1\}^{P^3}$ is constructed with a ratio of 75\% and the $L_2$ distance loss is used in the MAE loss:
\begin{equation}
    \mathcal{L}_{\text{baseline}} = \frac{1}{|\mathcal{M}|} \sum_{i \in \mathcal{M}} \left\| \hat{x}_i - x_i \right\|_2^2,
\end{equation}
where $\hat{x}=D\left(E\left((1-\mathcal{M}) \cdot x\right)\right)$ is the reconstruction of the patch $x$, masked by $\mathcal{M}$, by the encoder $E$-decoder $D$ architecture.
During pre-training, we also augment the data with random flips and orientations.

\paragraph{Choice of distance loss.} We challenge the choice of loss distance ($L_2$) in the context of medical imaging. Because $L_2$ assigns unbounded gradients, the small population of high-intensity outlier voxels characteristic of CT and MRI (e.g. bone, implants, contrast) dominates optimization. Thus, we choose to use instead the \textbf{Huber} (also known as \textbf{Smooth $L_1$}) \textbf{loss}, that caps the gradient of residuals beyond a given threshold $\beta$ while reallocating signal toward the bulk of small-to-moderate residuals that carry anatomical structure. Note that we use Smooth $L_1$ rather than pure $L_1$ because the latter has constant-magnitude gradients and a non-differentiable point at zero, which prevents stable refinement of well-predicted voxels. With this distance loss, the MAE loss becomes:
\begin{equation}
    \mathcal{L}_{\text{MAE}} = \frac{1}{|\mathcal{M}|} \sum_{i \in \mathcal{M}} L_\text{Huber}\left( \hat{x}_i - x_i \right),
\end{equation}
where the Huber loss is computed as:
\begin{equation}
    L_\text{Huber}(x)=
\begin{cases}
\frac{1}{2\beta}x^2 & \text{if } |x| < \beta, \\
|x| - \frac{1}{2}\beta & \text{if } |x| \ge \beta,
\end{cases}
\end{equation}
with a $\beta$ set to 0.3.

\paragraph{Context loss.} To supplement the standard MAE objective, we introduce a \textbf{context loss} $\mathcal{L}_\text{ctx}$, which functions as an additional reconstruction loss applied exclusively to the unmasked patches.
Without it, the visible regions are left unconstrained and need not be reconstructed faithfully, which can produce discontinuities at the boundary between visible and masked regions; the context loss acts as a light regularizer encouraging consistent reconstruction across this boundary. We weight it with a small $\lambda_\text{ctx}$ so that it supplements, rather than competes with, the primary masked-reconstruction objective.
%By penalizing reconstruction errors on the visible regions, this loss enforces a dual representation learning constraint: while the primary MAE loss drives the encoder to infer latent contextual features to predict the masked regions, the context loss ensures the preservation of high-fidelity features necessary to accurately reconstruct the originally observed anatomical structures. 
This loss is structurally similar to the MAE loss, with $\lambda_\text{ctx}=0.2$:
\begin{equation}
\mathcal{L}_{\text{ctx}} = \frac{\lambda_\text{ctx}}{P^3-|\mathcal{M}|} \sum_{i \notin \mathcal{M}} L_\text{Huber}\left( \hat{x}_i - x_i \right).
\end{equation}
 %balancing the impact of this context loss compared to the MAE loss which, in practice, we set to a small value of $0.2$.

\subsubsection{Regularization.}
We further regularize the encoder's feature space with a term inspired by \textbf{SIGReg}~\cite{balestriero2025lejepa} (Sketched Isotropic Gaussian Regularization), which encourages the distribution of embeddings toward an isotropic Gaussian $\mathcal{N}(0, I)$. Whereas the reconstruction objective places no constraint on feature geometry and may drift toward anisotropic or low-variance solutions that reconstruct well yet transfer poorly, an isotropic Gaussian target is intended to keep the latent space expressive and to prevent collapse to a degenerate low-dimensional subspace; under the analysis of \cite{balestriero2025lejepa}, it is the embedding distribution that minimizes worst-case downstream prediction risk.
%an isotropic Gaussian target keeps the latent space expressive, prevents collapse to a degenerate low-dimensional subspace, and minimizes worst-case downstream prediction risk across tasks~\cite{balestriero2025lejepa}. 
We apply the regularizer to the set of unmasked patch features ${z_i}$ collected over the batch, rather than to a class token, so that it shapes the dense representations actually consumed by downstream segmentation:
\begin{equation}
\mathcal{L}_{\text{reg}} = \lambda_{\text{reg}} \, \operatorname{SIGReg}({z_i}),
\end{equation}
where $\operatorname{SIGReg}(\cdot)$ matches the distribution of random 1D projections of ${z_i}$ to a standard Gaussian via a goodness-of-fit statistic; we refer to~\cite{balestriero2025lejepa} for details. We set $\lambda_{\text{reg}} = 0.1$ to keep the impact of the regularizer bounded.

\subsubsection{Random Spacing.} Volumetric medical images are acquired across a wide range of voxel spacings, owing to differences in scanners, acquisition protocols, and slice thickness. Resampling every volume to the fixed target spacing of $(1.0,1.0,1.0)$ exposes the encoder to a single resolution, leaving the learned representation brittle to the resolution shifts it encounters across sites and modalities at deployment. To build in robustness to this acquisition heterogeneity, we replace the fixed resampling target with a per-volume spacing sampled between the image's native spacing $s_{\text{orig}}$ and the canonical reference $s_{\text{ref}} = (1.0,1.0,1.0)$:
\begin{equation}
s = \alpha \, s_{\text{orig}} + (1-\alpha)\, s_{\text{ref}}, \qquad \alpha \sim \mathcal{U}(0,1).
\end{equation}
Anchoring the interval at the image's own native spacing on one end and the canonical target on the other makes the augmentation data-adaptive: each volume is perturbed only within the resolution range spanned by its raw acquisition and the standard pre-processing, rather than a fixed global range that could over- or under-resolve a given scan. Because the input patch size remains fixed at $P^3=160^3$, varying the spacing further induces a variation in physical field of view, exposing the encoder to anatomy across a range of scales. 
While we don't expect to observe direct benefits in our evaluation setting, as we will keep a fixed target spacing of $(1.0,1.0,1.0)$ for all tasks, robustness to random spacing is a desirable property for the training's global branch, described in the following section.

\subsubsection{Local--Global Similarity Objective.}
The reconstruction terms constrain the fidelity of the decoded volume and the SIGReg term constrains the marginal geometry of the feature distribution, but neither enforces that semantically equivalent inputs yield consistent representations: a feature space can be high-fidelity and isotropic yet still fail to place matching anatomy nearby in a linearly decodable way. To further constrain the features for transfer, we add a \textbf{local--global similarity} objective that aligns the representation of masked anatomy with the representation of the same region observed in the complete, unmasked volume. This directly targets the frozen-encoder downstream setup, in which reconstruction-only features are known to transfer poorly.

As illustrated in Fig.~\ref{fig:method}, we form two views of each volume. The local view crops a sub-region and applies the MAE masking; the global view only resizes the entire uncropped, unmasked volume to the input size. Both are passed through the shared encoder $E$. On the local branch, the features of the unmasked patches are pooled and projected into an embedding $z_l$. On the global branch, the features covering the spatial region corresponding to the crop are pooled by a weighted average (accounting for the region's fractional overlap with the coarser feature grid) and projected into a target embedding $z_g$. Aligning $z_l$ with $z_g$ thus asks the encoder to predict, in feature space, the representation of the local crop anatomy from a partial and corrupted view, using as target the features of the same region seen in the full volume. The loss thus enforces consistency both across corruption (masked versus complete) and across scale (the local crop versus the resized global volume).

Over a batch of matched pairs, the objective takes the general form
\begin{equation}
\mathcal{L}_{\text{sim}} = \lambda_\text{sim} \, \mathcal{D}\!\left(z_l, z_g\right),
\end{equation}
where $\mathcal{D}$ is a similarity criterion minimized when matched embeddings agree while preventing representational collapse. We set $\lambda_\text{sim} = 1.0$. We study two instantiations: \textbf{BYOL}~\cite{grill2020byol}, where $\mathcal{D}$ is a predictor-based negative cosine similarity with a stop-gradient on the target branch; and \textbf{Barlow Twins}~\cite{zbontar2021barlow}, where $\mathcal{D}$ drives the cross-correlation matrix between the two embeddings toward the identity, decorrelating feature dimensions to prevent collapse. Note that, unlike BYOL, Barlow Twins has no stop-gradient. We refer the reader to the original works for the detailed formulations and select empirically \textbf{Barlow Twins} for our final model (Section~\ref{sec:ablation_objective}).

\subsubsection{Final Objective.} With all the aforementioned changes considered, the final objective of our pre-training strategy can be written as:
\begin{equation}
\mathcal{L}_\text{tot}=\mathcal{L}_\text{MAE}+\mathcal{L}_\text{ctx}+\mathcal{L}_\text{reg}+\mathcal{L}_\text{sim}.
\end{equation}

\section{Experiments}
\label{sec:experiments}
\subsection{Pre-Training Setup}

The pre-training dataset, shared with \cite{dancette2025curia, saporta2026curia}, consists of routine cross-sectional clinical examinations acquired from 2019 to 2022 through the collaboration with a private hospital. The entire dataset was fully anonymized, which included the removal of identifying metadata and the defacing of scans containing the head region. After filtering based on volume shape, the final pre-training dataset contains 300,000 volumes across both CT and MRI modalities and spanning a wide range of anatomical regions across the body, such as the thorax, abdomen, pelvis, and head, to name a few.

We pre-train ResEnc-L~\cite{isensee2024nnu} networks using the AdamW optimizer with an initial learning rate of $0.01$. The per-GPU batch size was set to the largest value fitting in GPU memory: 12 with the local branch only, and 7 with both the global and local branches. To pre-train the final models, training was distributed across 16 nodes of the Leonardo supercomputer, with each node equipped with 4 NVIDIA A100 GPUs. For ablation experiments, 4 nodes were used.
Finally, note that we did not tune the loss weights or $\beta$ on the downstream benchmarks: $\beta=0.3$, $\lambda_\text{ctx}=0.2$, $\lambda_\text{reg}=0.1$, and $\lambda_\text{sim}=1.0$ were fixed to round default values prior to evaluation and kept fixed across all experiments.

% \textit{\noindent Prior to training, all volumetric images underwent intensity standardization via z-score normalization. The data augmentation pipeline comprising random orientation, random spacing \textbf{\textit{(do we want to develop more about that?)}}, and random cropping to a uniform target volume of $160 \times 160 \times 160$ voxels. }

\subsection{Segmentation Datasets}
To evaluate the performance of our pre-trained models, we assess them on downstream segmentation datasets across different modalities, namely CT and MRI. These datasets can be divided into two categories. 

First, an anatomy-focused benchmark which includes segmentation of general anatomical structures:
\begin{itemize}
    \item \textbf{AMOS-CT 2022 \cite{amos}}: A benchmark dataset containing annotations for 15 abdominal organs, widely recognized as a gold standard for multi-organ segmentation in CT imaging. \textit{Original splits (train/val: 200/100).}
    \item \textbf{AMOS-MRI 2022 \cite{amos}}: An MRI-focused extension of the AMOS-CT dataset, featuring the same 15 abdominal organs, thereby enabling cross-modality evaluation. \textit{Original splits (train/val: 40/20).}
    \item \textbf{TotalSegmentator-v2 CT (TS-CT) \cite{totalsegmentatorv2}}: A highly heterogeneous dataset encompassing 117 anatomical structures. It incorporates a wide range of CT images from diverse scanners and institutions, making it one of the most anatomically varied multi-organ benchmarks available. \textit{Original splits (train/val: 1082/57).}
    \item \textbf{TotalSegmentator-v2 MRI (TS-MRI) \cite{totalsegmentator_mri}}: A dataset comprising 50 anatomical structures delineated on MR images that were randomly sampled to capture a broad distribution of clinical MRI sequences. \textit{Original splits (train/val: 561/55).}
\end{itemize}

Second, a lesion-focused benchmark which includes segmentation of lesions in different anatomical sites:
\begin{itemize}
    \item \textbf{MSD-Liver \cite{msd}}: A CT dataset from the Medical Segmentation Decathlon (MSD) containing 2 classes: liver and liver lesions. \textit{Our splits (train/val: 91/40).}
    \item \textbf{MSD-Lung \cite{msd}}: A CT dataset from the MSD containing 1 class: lung lesions. \textit{Our splits (train/val: 44/19).}
    \item \textbf{MSD-Pancreas \cite{msd}}: A CT dataset from the MSD containing 2 classes: pancreas and pancreas lesions. \textit{Our splits (train/val: 196/85).}
    \item \textbf{ATLAS v2 \cite{atlasv2}}: A specialized dataset consisting of T1-weighted (T1w) MR images annotated for a single class corresponding to lesions after stroke. \textit{Our splits (train/val: 463/92).}
\end{itemize}

This evaluation framework, encompassing both diverse anatomical structures and heterogeneous pathological lesions across anatomical sites and modalities, is key to effectively validate the medical features captured during pre-training. 

\subsection{Segmentation Training and Evaluation}
For all segmentation fine-tuning experiments, the computational hardware consisted of a single node with 4 NVIDIA A100 GPUs, utilizing SGD as the optimizer. Following \cite{revisitingmae, nnunet}, models are trained for an equivalent of 1,000 epochs in the nnU-Net framework, leading to 250,000 optimization steps in our evaluation setup. All pre-trained models are evaluated in two setups: with the encoder kept frozen (shown with \textcolor{cyan!70}{\faSnowflake[regular]}  in tables); with the full model being fine-tuned (shown with \textcolor{orange!70}{\faFire} in tables). For both setups, we implemented warm-up scheduling strategies for both the encoder and decoder—following the methodology proposed in \cite{revisitingmae}—with a base learning rate of $2e^{-2}$.

Model performance is evaluated using the average Dice score over all classes of the dataset. The reported standard deviation is a bootstrap confidence interval over test cases (1000 resamples) and reflects evaluation uncertainty for a single trained model; it does not capture fine-tuning seed variance, as each configuration corresponds to a single pre-training run. Differences smaller than these intervals should be regarded as indicative rather than conclusive.

\subsection{Ablation Study}
\label{sec:ablation}
The ablation is split in two sections: we evaluate first the contributions of each component added to the local branch described in Section~\ref{sec:method}; then, we study the impact of the global branch and the local--global similarity objectives.

\subsubsection{Component Ablation.}
\begin{table}[t]
\centering
\caption{\textbf{Component ablation, anatomy-focused benchmark.} Each row cumulatively adds one component over the Baseline (MAE); Dice (\%) in the frozen-encoder (\textcolor{cyan!70}{\faSnowflake[regular]}) and
fine-tuned (\textcolor{orange!70}{\faFire}) regimes.}
\label{tab:c_results_anat}
\resizebox{\ifdim\width>\linewidth \linewidth\else \width\fi}{!}{%
\begin{tabular}{l|r@{}lr@{}l|r@{}lr@{}lr@{}lr@{}lr@{}lr@{}lr@{}lr@{}l}
\toprule
 & \multicolumn{4}{c|}{\textbf{Average}} & \multicolumn{4}{c}{AMOS-CT} & \multicolumn{4}{c}{AMOS-MRI} & \multicolumn{4}{c}{TS-CT} & \multicolumn{4}{c}{TS-MRI} \\
\cmidrule(lr){2-5} \cmidrule(lr){6-9} \cmidrule(lr){10-13} \cmidrule(lr){14-17} \cmidrule(lr){18-21}
\textbf{Encoder State} & \multicolumn{2}{c}{\small\textcolor{cyan!70}{\faSnowflake[regular]}} & \multicolumn{2}{c|}{\small\textcolor{orange!70}{\faFire}} & \multicolumn{2}{c}{\small\textcolor{cyan!70}{\faSnowflake[regular]}} & \multicolumn{2}{c}{\small\textcolor{orange!70}{\faFire}} & \multicolumn{2}{c}{\small\textcolor{cyan!70}{\faSnowflake[regular]}} & \multicolumn{2}{c}{\small\textcolor{orange!70}{\faFire}} & \multicolumn{2}{c}{\small\textcolor{cyan!70}{\faSnowflake[regular]}} & \multicolumn{2}{c}{\small\textcolor{orange!70}{\faFire}} & \multicolumn{2}{c}{\small\textcolor{cyan!70}{\faSnowflake[regular]}} & \multicolumn{2}{c}{\small\textcolor{orange!70}{\faFire}} \\
\midrule
Baseline (MAE) & 64.4 & \,${\scriptstyle \pm 1.3}$ & 79.1 & \,${\scriptstyle \pm 1.3}$ & 79.4 & \,${\scriptstyle \pm 0.7}$ & 88.2 & \,${\scriptstyle \pm 0.3}$ & 81.1 & \,${\scriptstyle \pm 1.3}$ & 83.6 & \,${\scriptstyle \pm 1.3}$ & 62.8 & \,${\scriptstyle \pm 1.1}$ & 82.3 & \,${\scriptstyle \pm 1.0}$ & 34.4 & \,${\scriptstyle \pm 2.0}$ & 62.4 & \,${\scriptstyle \pm 2.6}$ \\
$+$ Smooth $L_1$ & 65.0 & \,${\scriptstyle \pm 1.3}$ & \textbf{79.9} & \,${\scriptstyle \pm 1.2}$ & 81.1 & \,${\scriptstyle \pm 0.6}$ & \textbf{88.8} & \,${\scriptstyle \pm 0.3}$ & 81.2 & \,${\scriptstyle \pm 1.3}$ & 84.4 & \,${\scriptstyle \pm 1.2}$ & 62.9 & \,${\scriptstyle \pm 1.3}$ & \textbf{82.9} & \,${\scriptstyle \pm 1.0}$ & 34.9 & \,${\scriptstyle \pm 1.9}$ & \textbf{63.6} & \,${\scriptstyle \pm 2.2}$ \\
$+$ Context & \textbf{65.8} & \,${\scriptstyle \pm 1.3}$ & 78.8 & \,${\scriptstyle \pm 1.2}$ & \textbf{82.4} & \,${\scriptstyle \pm 0.6}$ & 88.0 & \,${\scriptstyle \pm 0.3}$ & \textbf{82.0} & \,${\scriptstyle \pm 1.2}$ & 84.0 & \,${\scriptstyle \pm 1.1}$ & 62.9 & \,${\scriptstyle \pm 1.2}$ & 80.9 & \,${\scriptstyle \pm 1.1}$ & \textbf{36.0} & \,${\scriptstyle \pm 2.2}$ & 62.4 & \,${\scriptstyle \pm 2.4}$ \\
$+$ Reg & 64.6 & \,${\scriptstyle \pm 1.3}$ & 79.7 & \,${\scriptstyle \pm 1.2}$ & 80.8 & \,${\scriptstyle \pm 0.6}$ & 88.5 & \,${\scriptstyle \pm 0.3}$ & 80.8 & \,${\scriptstyle \pm 1.4}$ & \textbf{84.7} & \,${\scriptstyle \pm 1.0}$ & 61.1 & \,${\scriptstyle \pm 1.2}$ & 81.9 & \,${\scriptstyle \pm 1.1}$ & 35.8 & \,${\scriptstyle \pm 2.1}$ & \textbf{63.6} & \,${\scriptstyle \pm 2.5}$ \\
$+$ Spacing & 65.1 & \,${\scriptstyle \pm 1.3}$ & 79.3 & \,${\scriptstyle \pm 1.3}$ & 81.4 & \,${\scriptstyle \pm 0.6}$ & 88.5 & \,${\scriptstyle \pm 0.3}$ & 80.0 & \,${\scriptstyle \pm 1.4}$ & 83.9 & \,${\scriptstyle \pm 1.1}$ & \textbf{63.5} & \,${\scriptstyle \pm 1.1}$ & 82.3 & \,${\scriptstyle \pm 1.1}$ & 35.6 & \,${\scriptstyle \pm 2.0}$ & 62.7 & \,${\scriptstyle \pm 2.5}$ \\
\bottomrule
\end{tabular}
}
\end{table}

\begin{table}[t]
\centering
\caption{\textbf{Component ablation, lesion-focused benchmark.} Each row cumulatively adds one component over the Baseline (MAE); Dice (\%) in the frozen-encoder (\textcolor{cyan!70}{\faSnowflake[regular]}) and
fine-tuned (\textcolor{orange!70}{\faFire}) regimes.}
\label{tab:c_results_lesion}
\resizebox{\ifdim\width>\linewidth \linewidth\else \width\fi}{!}{%
\begin{tabular}{l|r@{}lr@{}l|r@{}lr@{}lr@{}lr@{}lr@{}lr@{}lr@{}lr@{}l}
\toprule
 & \multicolumn{4}{c|}{\textbf{Average}} & \multicolumn{4}{c}{MSD-Liver} & \multicolumn{4}{c}{MSD-Lung} & \multicolumn{4}{c}{MSD-Pancreas} & \multicolumn{4}{c}{Atlas} \\
\cmidrule(lr){2-5} \cmidrule(lr){6-9} \cmidrule(lr){10-13} \cmidrule(lr){14-17} \cmidrule(lr){18-21}
\textbf{Encoder State} & \multicolumn{2}{c}{\small\textcolor{cyan!70}{\faSnowflake[regular]}} & \multicolumn{2}{c|}{\small\textcolor{orange!70}{\faFire}} & \multicolumn{2}{c}{\small\textcolor{cyan!70}{\faSnowflake[regular]}} & \multicolumn{2}{c}{\small\textcolor{orange!70}{\faFire}} & \multicolumn{2}{c}{\small\textcolor{cyan!70}{\faSnowflake[regular]}} & \multicolumn{2}{c}{\small\textcolor{orange!70}{\faFire}} & \multicolumn{2}{c}{\small\textcolor{cyan!70}{\faSnowflake[regular]}} & \multicolumn{2}{c}{\small\textcolor{orange!70}{\faFire}} & \multicolumn{2}{c}{\small\textcolor{cyan!70}{\faSnowflake[regular]}} & \multicolumn{2}{c}{\small\textcolor{orange!70}{\faFire}} \\
\midrule
Baseline (MAE) & 61.9 & \,${\scriptstyle \pm 2.9}$ & 69.2 & \,${\scriptstyle \pm 2.9}$ & 70.5 & \,${\scriptstyle \pm 2.5}$ & 80.0 & \,${\scriptstyle \pm 2.4}$ & 69.6 & \,${\scriptstyle \pm 4.1}$ & 71.4 & \,${\scriptstyle \pm 4.5}$ & 51.9 & \,${\scriptstyle \pm 1.7}$ & 66.9 & \,${\scriptstyle \pm 1.9}$ & \textbf{55.7} & \,${\scriptstyle \pm 3.1}$ & 58.5 & \,${\scriptstyle \pm 3.0}$ \\
$+$ Smooth $L_1$ & \textbf{63.0} & \,${\scriptstyle \pm 2.9}$ & 68.4 & \,${\scriptstyle \pm 3.1}$ & 70.9 & \,${\scriptstyle \pm 2.6}$ & 81.0 & \,${\scriptstyle \pm 2.2}$ & \textbf{72.3} & \,${\scriptstyle \pm 4.3}$ & 65.8 & \,${\scriptstyle \pm 5.7}$ & 53.7 & \,${\scriptstyle \pm 1.8}$ & 67.4 & \,${\scriptstyle \pm 1.8}$ & 54.9 & \,${\scriptstyle \pm 3.1}$ & \textbf{59.4} & \,${\scriptstyle \pm 2.8}$ \\
$+$ Context & 61.0 & \,${\scriptstyle \pm 3.4}$ & 68.9 & \,${\scriptstyle \pm 2.8}$ & 71.1 & \,${\scriptstyle \pm 2.7}$ & 81.4 & \,${\scriptstyle \pm 2.2}$ & 64.2 & \,${\scriptstyle \pm 5.8}$ & 70.7 & \,${\scriptstyle \pm 4.4}$ & \textbf{54.4} & \,${\scriptstyle \pm 1.9}$ & 66.0 & \,${\scriptstyle \pm 1.8}$ & 54.4 & \,${\scriptstyle \pm 3.2}$ & 57.6 & \,${\scriptstyle \pm 2.9}$ \\
$+$ Reg & 61.4 & \,${\scriptstyle \pm 3.0}$ & \textbf{70.2} & \,${\scriptstyle \pm 2.6}$ & 70.9 & \,${\scriptstyle \pm 2.6}$ & 80.8 & \,${\scriptstyle \pm 2.2}$ & 70.5 & \,${\scriptstyle \pm 4.3}$ & \textbf{73.9} & \,${\scriptstyle \pm 3.6}$ & 48.8 & \,${\scriptstyle \pm 1.9}$ & 67.9 & \,${\scriptstyle \pm 1.8}$ & 55.4 & \,${\scriptstyle \pm 3.0}$ & 58.2 & \,${\scriptstyle \pm 3.0}$ \\
$+$ Spacing & 60.8 & \,${\scriptstyle \pm 3.1}$ & 69.2 & \,${\scriptstyle \pm 3.1}$ & \textbf{71.3} & \,${\scriptstyle \pm 2.5}$ & \textbf{81.6} & \,${\scriptstyle \pm 2.1}$ & 66.8 & \,${\scriptstyle \pm 4.8}$ & 70.2 & \,${\scriptstyle \pm 5.3}$ & 50.8 & \,${\scriptstyle \pm 1.9}$ & \textbf{68.0} & \,${\scriptstyle \pm 1.8}$ & 54.5 & \,${\scriptstyle \pm 3.1}$ & 57.0 & \,${\scriptstyle \pm 3.2}$ \\
\bottomrule
\end{tabular}
}
\end{table}
We ablate each component cumulatively on the per-dataset anatomical (Table~\ref{tab:c_results_anat}) and lesion (Table~\ref{tab:c_results_lesion}) benchmarks, reporting Dice in both the frozen-encoder and fine-tuned regimes. Each component is designed to target a specific axis (reconstruction fidelity, anatomical consistency, feature geometry, or acquisition robustness) and we analyze its effect across both anatomical and lesion tasks accordingly.

We analyze each component on the axis it targets, and report effects plainly, including where they are neutral or negative on the headline frozen-encoder metric. We emphasize that the frozen-encoder gains of the full method are driven primarily by the local--global objective (Section~\ref{sec:ablation_objective}); the reconstruction-side components are retained for their effect on fine-tuning robustness and acquisition robustness rather than for large standalone frozen-encoder gains.

Smooth $L_1$ is the only reconstruction component to improve the frozen-encoder average on both task families (anatomical $64.4\rightarrow65.0$, lesion $61.9\rightarrow63.0$), consistent with reducing the influence of high-intensity outlier voxels; it also improves the anatomical fine-tuned average. The context loss improves anatomical frozen-encoder performance (to $65.8$, its best in the build-up) but is neutral-to-negative on lesion-focused tasks, matching its organ-centric motivation. The feature regularizer is fine-tuning oriented: it improves fine-tuned averages while slightly reducing frozen-encoder performance on individual tasks. Random spacing leaves in-distribution Dice essentially unchanged by design, targeting resolution robustness rather than benchmark accuracy. Note in particular that robustness to random spacing is a desirable property for the global branch, ablated in the next section, since the global volume is reshaped to the encoder's input shape without any spacing consideration. Cumulatively, these components maintain fine-tuning performance and add reconstruction and acquisition robustness, while the principal frozen-encoder improvement comes from the local--global objective studied next.

\subsubsection{Local--Global Similarity Ablation.}
\label{sec:ablation_objective}
\begin{table}[t]
\centering
\caption{\textbf{Local--global similarity ablation, anatomy-focused benchmark.} BYOL and Barlow Twins are added on top of all components from Table~\ref{tab:c_results_anat} (model \O); Dice (\%) in the frozen-encoder (\textcolor{cyan!70}{\faSnowflake[regular]}) and
fine-tuned (\textcolor{orange!70}{\faFire}) regimes.}
\label{tab:o_results_anat}
\resizebox{\ifdim\width>\linewidth \linewidth\else \width\fi}{!}{%
\begin{tabular}{l|r@{}lr@{}l|r@{}lr@{}lr@{}lr@{}lr@{}lr@{}lr@{}lr@{}l}
\toprule
 & \multicolumn{4}{c|}{\textbf{Average}} & \multicolumn{4}{c}{AMOS-CT} & \multicolumn{4}{c}{AMOS-MRI} & \multicolumn{4}{c}{TS-CT} & \multicolumn{4}{c}{TS-MRI} \\
\cmidrule(lr){2-5} \cmidrule(lr){6-9} \cmidrule(lr){10-13} \cmidrule(lr){14-17} \cmidrule(lr){18-21}
\textbf{Encoder State} & \multicolumn{2}{c}{\small\textcolor{cyan!70}{\faSnowflake[regular]}} & \multicolumn{2}{c|}{\small\textcolor{orange!70}{\faFire}} & \multicolumn{2}{c}{\small\textcolor{cyan!70}{\faSnowflake[regular]}} & \multicolumn{2}{c}{\small\textcolor{orange!70}{\faFire}} & \multicolumn{2}{c}{\small\textcolor{cyan!70}{\faSnowflake[regular]}} & \multicolumn{2}{c}{\small\textcolor{orange!70}{\faFire}} & \multicolumn{2}{c}{\small\textcolor{cyan!70}{\faSnowflake[regular]}} & \multicolumn{2}{c}{\small\textcolor{orange!70}{\faFire}} & \multicolumn{2}{c}{\small\textcolor{cyan!70}{\faSnowflake[regular]}} & \multicolumn{2}{c}{\small\textcolor{orange!70}{\faFire}} \\
\midrule
Baseline (MAE) & 64.4 & \,${\scriptstyle \pm 1.3}$ & 79.1 & \,${\scriptstyle \pm 1.3}$ & 79.4 & \,${\scriptstyle \pm 0.7}$ & 88.2 & \,${\scriptstyle \pm 0.3}$ & 81.1 & \,${\scriptstyle \pm 1.3}$ & 83.6 & \,${\scriptstyle \pm 1.3}$ & 62.8 & \,${\scriptstyle \pm 1.1}$ & 82.3 & \,${\scriptstyle \pm 1.0}$ & 34.4 & \,${\scriptstyle \pm 2.0}$ & 62.4 & \,${\scriptstyle \pm 2.6}$ \\
\midrule
\O & 65.1 & \,${\scriptstyle \pm 1.3}$ & \textbf{79.3} & \,${\scriptstyle \pm 1.3}$ & 81.4 & \,${\scriptstyle \pm 0.6}$ & \textbf{88.5} & \,${\scriptstyle \pm 0.3}$ & 80.0 & \,${\scriptstyle \pm 1.4}$ & \textbf{83.9} & \,${\scriptstyle \pm 1.1}$ & 63.5 & \,${\scriptstyle \pm 1.1}$ & 82.3 & \,${\scriptstyle \pm 1.1}$ & 35.6 & \,${\scriptstyle \pm 2.0}$ & \textbf{62.7} & \,${\scriptstyle \pm 2.5}$ \\
BYOL & 61.1 & \,${\scriptstyle \pm 1.4}$ & 77.6 & \,${\scriptstyle \pm 1.3}$ & 78.4 & \,${\scriptstyle \pm 0.7}$ & 88.1 & \,${\scriptstyle \pm 0.3}$ & 77.6 & \,${\scriptstyle \pm 1.8}$ & 83.6 & \,${\scriptstyle \pm 1.2}$ & 58.0 & \,${\scriptstyle \pm 1.3}$ & 81.5 & \,${\scriptstyle \pm 1.1}$ & 30.3 & \,${\scriptstyle \pm 1.9}$ & 57.4 & \,${\scriptstyle \pm 2.5}$ \\
Barlow Twins & \textbf{66.3} & \,${\scriptstyle \pm 1.2}$ & 78.9 & \,${\scriptstyle \pm 1.3}$ & \textbf{82.1} & \,${\scriptstyle \pm 0.6}$ & 88.3 & \,${\scriptstyle \pm 0.3}$ & \textbf{81.3} & \,${\scriptstyle \pm 1.2}$ & 83.5 & \,${\scriptstyle \pm 1.3}$ & \textbf{65.8} & \,${\scriptstyle \pm 1.2}$ & \textbf{82.7} & \,${\scriptstyle \pm 1.1}$ & \textbf{36.1} & \,${\scriptstyle \pm 2.0}$ & 61.1 & \,${\scriptstyle \pm 2.5}$ \\
\bottomrule
\end{tabular}
}
\end{table}

\begin{table}[t]
\centering
\caption{\textbf{Local-global similarity ablation, lesion-focused benchmark.} BYOL and Barlow Twins are added on top of all components from Table~\ref{tab:c_results_lesion} (model \O); Dice (\%) in the frozen-encoder (\textcolor{cyan!70}{\faSnowflake[regular]}) and
fine-tuned (\textcolor{orange!70}{\faFire}) regimes.}
\label{tab:o_results_lesion}
\resizebox{\ifdim\width>\linewidth \linewidth\else \width\fi}{!}{%
\begin{tabular}{l|r@{}lr@{}l|r@{}lr@{}lr@{}lr@{}lr@{}lr@{}lr@{}lr@{}l}
\toprule
 & \multicolumn{4}{c|}{\textbf{Average}} & \multicolumn{4}{c}{MSD-Liver} & \multicolumn{4}{c}{MSD-Lung} & \multicolumn{4}{c}{MSD-Pancreas} & \multicolumn{4}{c}{Atlas} \\
\cmidrule(lr){2-5} \cmidrule(lr){6-9} \cmidrule(lr){10-13} \cmidrule(lr){14-17} \cmidrule(lr){18-21}
\textbf{Encoder State} & \multicolumn{2}{c}{\small\textcolor{cyan!70}{\faSnowflake[regular]}} & \multicolumn{2}{c|}{\small\textcolor{orange!70}{\faFire}} & \multicolumn{2}{c}{\small\textcolor{cyan!70}{\faSnowflake[regular]}} & \multicolumn{2}{c}{\small\textcolor{orange!70}{\faFire}} & \multicolumn{2}{c}{\small\textcolor{cyan!70}{\faSnowflake[regular]}} & \multicolumn{2}{c}{\small\textcolor{orange!70}{\faFire}} & \multicolumn{2}{c}{\small\textcolor{cyan!70}{\faSnowflake[regular]}} & \multicolumn{2}{c}{\small\textcolor{orange!70}{\faFire}} & \multicolumn{2}{c}{\small\textcolor{cyan!70}{\faSnowflake[regular]}} & \multicolumn{2}{c}{\small\textcolor{orange!70}{\faFire}} \\
\midrule
Baseline (MAE) & 61.9 & \,${\scriptstyle \pm 2.9}$ & \textbf{69.2} & \,${\scriptstyle \pm 2.9}$ & 70.5 & \,${\scriptstyle \pm 2.5}$ & 80.0 & \,${\scriptstyle \pm 2.4}$ & \textbf{69.6} & \,${\scriptstyle \pm 4.1}$ & \textbf{71.4} & \,${\scriptstyle \pm 4.5}$ & 51.9 & \,${\scriptstyle \pm 1.7}$ & 66.9 & \,${\scriptstyle \pm 1.9}$ & 55.7 & \,${\scriptstyle \pm 3.1}$ & \textbf{58.5} & \,${\scriptstyle \pm 3.0}$ \\
\midrule
\O & 60.8 & \,${\scriptstyle \pm 3.1}$ & \textbf{69.2} & \,${\scriptstyle \pm 3.1}$ & 71.3 & \,${\scriptstyle \pm 2.5}$ & \textbf{81.6} & \,${\scriptstyle \pm 2.1}$ & 66.8 & \,${\scriptstyle \pm 4.8}$ & 70.2 & \,${\scriptstyle \pm 5.3}$ & 50.8 & \,${\scriptstyle \pm 1.9}$ & \textbf{68.0} & \,${\scriptstyle \pm 1.8}$ & 54.5 & \,${\scriptstyle \pm 3.1}$ & 57.0 & \,${\scriptstyle \pm 3.2}$ \\
BYOL & 58.2 & \,${\scriptstyle \pm 3.4}$ & 67.8 & \,${\scriptstyle \pm 2.9}$ & 65.3 & \,${\scriptstyle \pm 2.8}$ & 80.3 & \,${\scriptstyle \pm 2.2}$ & 63.6 & \,${\scriptstyle \pm 5.7}$ & 67.1 & \,${\scriptstyle \pm 4.4}$ & 50.0 & \,${\scriptstyle \pm 1.9}$ & 67.4 & \,${\scriptstyle \pm 1.8}$ & 54.0 & \,${\scriptstyle \pm 3.2}$ & 56.5 & \,${\scriptstyle \pm 3.1}$ \\
Barlow Twins & \textbf{63.3} & \,${\scriptstyle \pm 3.1}$ & 67.5 & \,${\scriptstyle \pm 3.2}$ & \textbf{72.3} & \,${\scriptstyle \pm 2.6}$ & 80.8 & \,${\scriptstyle \pm 2.4}$ & 69.3 & \,${\scriptstyle \pm 4.8}$ & 65.7 & \,${\scriptstyle \pm 5.7}$ & \textbf{55.8} & \,${\scriptstyle \pm 1.9}$ & 66.9 & \,${\scriptstyle \pm 1.8}$ & \textbf{55.8} & \,${\scriptstyle \pm 3.1}$ & 56.8 & \,${\scriptstyle \pm 3.1}$ \\
\bottomrule
\end{tabular}
}
\end{table}
We compare the two instantiations of the local--global similarity objective against the full component stack without any joint-embedding term (\O), with the MAE baseline reported as a reference (Table~\ref{tab:o_results_anat}, Table~\ref{tab:o_results_lesion}). The objective is intended to constrain the feature space for transfer, so we read its effect primarily in the frozen-encoder regime.

Barlow Twins delivers the principal frozen-encoder gain of the method. On the anatomical benchmarks it reaches the best frozen-encoder average ($66.3$, against $65.1$ without the objective and $64.4$ for the baseline), with the strongest improvements on AMOS-CT, AMOS-MRI, and TS-CT; fine-tuned performance is preserved ($78.9$). On the lesion benchmarks the pattern is the same and more pronounced: it attains the best frozen-encoder average ($63.3$, up from $60.8$), with clear gains on MSD-Liver and MSD-Pancreas. Across both task families, adding the objective improves the intrinsic quality of the frozen representation while leaving fine-tuning essentially unchanged.

BYOL, by contrast, degrades both regimes on both task families: it lowers the frozen-encoder average on the anatomical ($61.1$) and lesion ($58.2$) benchmarks below not only Barlow Twins but the no-objective configuration and the baseline, with corresponding drops in fine-tuning. We attribute this to the instability of negative-free, predictor-based alignment in our small-batch 3D regime, where the absence of an explicit decorrelation mechanism admits partial representational collapse. The redundancy-reduction criterion of Barlow Twins avoids this failure mode, and we therefore adopt it as the joint-embedding objective in our method.%final method; BYOL is reported here only as an alternative we evaluated and discarded.

\subsection{Segmentation Results}
\begin{table}[t]
\centering
\caption{\textbf{Segmentation results, anatomy-focused benchmark.} \ours{} and Baseline (MAE) against nnU-Net and from-scratch ResEnc-L; Dice (\%) in the frozen-encoder (\textcolor{cyan!70}{\faSnowflake[regular]}) and
fine-tuned (\textcolor{orange!70}{\faFire}) regimes.}
\label{tab:results_anat}
\resizebox{\ifdim\width>\linewidth \linewidth\else \width\fi}{!}{%
\begin{tabular}{l|r@{}lr@{}l|r@{}lr@{}lr@{}lr@{}lr@{}lr@{}lr@{}lr@{}l}
\toprule
 & \multicolumn{4}{c|}{\textbf{Average}} & \multicolumn{4}{c}{AMOS-CT} & \multicolumn{4}{c}{AMOS-MRI} & \multicolumn{4}{c}{TS-CT} & \multicolumn{4}{c}{TS-MRI} \\
\cmidrule(lr){2-5} \cmidrule(lr){6-9} \cmidrule(lr){10-13} \cmidrule(lr){14-17} \cmidrule(lr){18-21}
\textbf{Encoder State} & \multicolumn{2}{c}{\small\textcolor{cyan!70}{\faSnowflake[regular]}} & \multicolumn{2}{c|}{\small\textcolor{orange!70}{\faFire}} & \multicolumn{2}{c}{\small\textcolor{cyan!70}{\faSnowflake[regular]}} & \multicolumn{2}{c}{\small\textcolor{orange!70}{\faFire}} & \multicolumn{2}{c}{\small\textcolor{cyan!70}{\faSnowflake[regular]}} & \multicolumn{2}{c}{\small\textcolor{orange!70}{\faFire}} & \multicolumn{2}{c}{\small\textcolor{cyan!70}{\faSnowflake[regular]}} & \multicolumn{2}{c}{\small\textcolor{orange!70}{\faFire}} & \multicolumn{2}{c}{\small\textcolor{cyan!70}{\faSnowflake[regular]}} & \multicolumn{2}{c}{\small\textcolor{orange!70}{\faFire}} \\
\midrule
nnU-Net (org) \cite{nnunet} & -- &  & \textbf{81.8} &  & -- &  & \textbf{90.2} &  & -- &  & \textbf{87.1} &  & -- &  & \textbf{86.0} &  & -- &  & \textbf{64.0} &  \\
ResEnc-L (scratch) & -- &  & 79.3 & \,${\scriptstyle \pm 1.3}$ & -- &  & 88.9 & \,${\scriptstyle \pm 0.3}$ & -- &  & 84.2 & \,${\scriptstyle \pm 1.2}$ & -- &  & 83.1 & \,${\scriptstyle \pm 1.0}$ & -- &  & 61.3 & \,${\scriptstyle \pm 2.5}$ \\
Baseline (MAE) & 66.8 & \,${\scriptstyle \pm 1.2}$ & 78.5 & \,${\scriptstyle \pm 1.2}$ & 81.3 & \,${\scriptstyle \pm 0.6}$ & 88.1 & \,${\scriptstyle \pm 0.3}$ & \textbf{82.1} & \,${\scriptstyle \pm 1.3}$ & 84.0 & \,${\scriptstyle \pm 1.2}$ & \textbf{67.0} & \,${\scriptstyle \pm 1.2}$ & 82.3 & \,${\scriptstyle \pm 1.1}$ & 36.9 & \,${\scriptstyle \pm 1.9}$ & 59.5 & \,${\scriptstyle \pm 2.2}$ \\
\ours & \textbf{67.5} & \,${\scriptstyle \pm 1.3}$ & 78.5 & \,${\scriptstyle \pm 1.2}$ & \textbf{83.3} & \,${\scriptstyle \pm 0.6}$ & 88.1 & \,${\scriptstyle \pm 0.4}$ & 81.0 & \,${\scriptstyle \pm 1.3}$ & 83.5 & \,${\scriptstyle \pm 1.1}$ & 65.2 & \,${\scriptstyle \pm 1.2}$ & 82.1 & \,${\scriptstyle \pm 1.1}$ & \textbf{40.6} & \,${\scriptstyle \pm 2.2}$ & 60.3 & \,${\scriptstyle \pm 2.2}$ \\
\bottomrule
\end{tabular}
}
\end{table}

\begin{table}[t]
\centering
\caption{\textbf{Segmentation results, lesion-focused benchmark.} \ours{} and Baseline (MAE) against nnU-Net and from-scratch ResEnc-L; Dice (\%) in the frozen-encoder (\textcolor{cyan!70}{\faSnowflake[regular]}) and
fine-tuned (\textcolor{orange!70}{\faFire}) regimes.}
\label{tab:results_lesion}
\resizebox{\ifdim\width>\linewidth \linewidth\else \width\fi}{!}{%
\begin{tabular}{l|r@{}lr@{}l|r@{}lr@{}lr@{}lr@{}lr@{}lr@{}lr@{}lr@{}l}
\toprule
 & \multicolumn{4}{c|}{\textbf{Average}} & \multicolumn{4}{c}{MSD-Liver} & \multicolumn{4}{c}{MSD-Lung} & \multicolumn{4}{c}{MSD-Pancreas} & \multicolumn{4}{c}{Atlas} \\
\cmidrule(lr){2-5} \cmidrule(lr){6-9} \cmidrule(lr){10-13} \cmidrule(lr){14-17} \cmidrule(lr){18-21}
\textbf{Encoder State} & \multicolumn{2}{c}{\small\textcolor{cyan!70}{\faSnowflake[regular]}} & \multicolumn{2}{c|}{\small\textcolor{orange!70}{\faFire}} & \multicolumn{2}{c}{\small\textcolor{cyan!70}{\faSnowflake[regular]}} & \multicolumn{2}{c}{\small\textcolor{orange!70}{\faFire}} & \multicolumn{2}{c}{\small\textcolor{cyan!70}{\faSnowflake[regular]}} & \multicolumn{2}{c}{\small\textcolor{orange!70}{\faFire}} & \multicolumn{2}{c}{\small\textcolor{cyan!70}{\faSnowflake[regular]}} & \multicolumn{2}{c}{\small\textcolor{orange!70}{\faFire}} & \multicolumn{2}{c}{\small\textcolor{cyan!70}{\faSnowflake[regular]}} & \multicolumn{2}{c}{\small\textcolor{orange!70}{\faFire}} \\
\midrule
nnU-Net (org) \cite{nnunet} & -- &  & 68.6 &  & -- &  & 79.6 &  & -- &  & \textbf{73.7} &  & -- &  & 62.0 &  & -- &  & \textbf{59.0} &  \\
ResEnc-L (scratch) & -- &  & 69.1 & \,${\scriptstyle \pm 3.0}$ & -- &  & 80.7 & \,${\scriptstyle \pm 2.3}$ & -- &  & 70.5 & \,${\scriptstyle \pm 5.2}$ & -- &  & 67.1 & \,${\scriptstyle \pm 1.9}$ & -- &  & 57.9 & \,${\scriptstyle \pm 2.8}$ \\
Baseline (MAE) & 63.0 & \,${\scriptstyle \pm 3.2}$ & 68.9 & \,${\scriptstyle \pm 2.5}$ & \textbf{73.8} & \,${\scriptstyle \pm 2.4}$ & 78.0 & \,${\scriptstyle \pm 2.3}$ & 68.6 & \,${\scriptstyle \pm 5.4}$ & 72.2 & \,${\scriptstyle \pm 3.0}$ & 55.2 & \,${\scriptstyle \pm 1.8}$ & 67.5 & \,${\scriptstyle \pm 1.8}$ & 54.4 & \,${\scriptstyle \pm 3.2}$ & 58.1 & \,${\scriptstyle \pm 3.0}$ \\
\ours & \textbf{64.7} & \,${\scriptstyle \pm 2.6}$ & \textbf{69.4} & \,${\scriptstyle \pm 2.6}$ & 71.0 & \,${\scriptstyle \pm 2.4}$ & \textbf{81.7} & \,${\scriptstyle \pm 2.1}$ & \textbf{73.4} & \,${\scriptstyle \pm 3.2}$ & 71.5 & \,${\scriptstyle \pm 3.7}$ & \textbf{57.9} & \,${\scriptstyle \pm 1.8}$ & \textbf{68.0} & \,${\scriptstyle \pm 1.6}$ & \textbf{56.6} & \,${\scriptstyle \pm 2.9}$ & 56.4 & \,${\scriptstyle \pm 3.1}$ \\
\bottomrule
\end{tabular}
}
\end{table}
Both the MAE baseline and \ours{} in this section are trained at the full
16-node scale (longer schedule, full pre-training set), whereas the ablations
(Section~\ref{sec:ablation}) use a 4-node regime. Absolute numbers therefore differ
from Tables~\ref{tab:c_results_anat}--\ref{tab:o_results_lesion} (e.g. the baseline anatomical frozen-encoder average is $64.4$ at ablation scale and $66.8$ at full scale); we retrain the baseline under identical full-scale conditions for a fair comparison.
We compare our final model \textbf{\ours{}} against two fully-supervised references trained from scratch on each downstream task: nnU-Net~\cite{nnunet}, state-of-the-art segmentation model trained with the official pipeline on our dataset splits, and ResEnc-L~\cite{isensee2024nnu}, which shares the encoder architecture and data pre-processing of our pre-trained models. The frozen encoder regime is reported only for the self-supervised models. We report both the anatomy-focused (Table~\ref{tab:results_anat}) and lesion-focused (Table~\ref{tab:results_lesion}) benchmarks.

In the frozen-encoder regime, \ours{} yields the best representation on both task families, improving the average over the MAE baseline on the anatomical ($66.8 \rightarrow 67.5$) and lesion ($63.0 \rightarrow 64.7$) benchmarks. At the dataset level the picture is mixed: \ours{} improves anatomical frozen-encoder performance on AMOS-CT ($81.3\rightarrow83.3$) and notably on TS-MRI ($36.9\rightarrow40.6$), while slightly declining on AMOS-MRI and TS-CT. The TS-MRI gain is the most pronounced, though the low absolute Dice ($\sim$40) indicates the frozen representation remains weak on that task. On lesion-focused tasks the average improvement is more uniform across datasets: 
%The gains are consistent at the dataset level: on anatomy, \ours{} reaches the best frozen-encoder score on AMOS-CT ($83.3$) and substantially improves the difficult TS-MRI task ($36.9 \rightarrow 40.6$); on lesions, 
\ours{} attains the best frozen-encoder score on three of four datasets (MSD-Lung, MSD-Pancreas, Atlas).
The practical relevance of these numbers lies in the frozen-encoder setting itself: with the encoder entirely frozen and only a decoder trained per task, \ours{} approaches its own fine-tuned performance on several lesion-focused tasks, indicating that the representation is already strongly task-relevant without any encoder adaptation.

Under full fine-tuning, the picture differs by task family. On the lesion-focused benchmarks, fine-tuning our pre-trained encoder gives the best fine-tuned average ($69.4$), exceeding both ResEnc-L ($69.1$) and nnU-Net ($68.6$), with the best fine-tuned scores on MSD-Liver and MSD-Pancreas; self-supervised pre-training therefore provides a useful initialization precisely on the smaller, harder lesion tasks where data is scarce. On the anatomy-focused benchmarks, by contrast, fine-tuning is competitive but does not surpass the supervised references ($78.5$ for \ours{}, against $79.3$ for ResEnc-L and $81.8$ for nnU-Net): a strong supervised model trained from scratch remains hard to beat, and pre-training does not help the fine-tuned result. Crucially, our additions deliver the frozen-encoder gains without degrading fine-tuned performance relative to the MAE baseline (anatomy: $78.5$ in both cases; lesion: $68.9 \rightarrow 69.4$).

Taken together, the results support the intended positioning of the method: it produces a markedly stronger frozen representation across whole-body, multi-modal anatomical and lesion segmentation, while remaining competitive, and, on lesions, superior, under full fine-tuning. The value of the approach is thus most pronounced wherever a single pre-trained encoder must be reused across tasks without per-task encoder training.

\subsection{Classification and Regression Results}
\begin{table}[t]
    \centering

    \caption{\textbf{Classification and regression on CuriaBench3D~\cite{saporta2026curia}.}
Frozen-encoder performance on classification, regression, and survival tasks; per-task metrics indicated in the header. All scores except \ours{} and the Baseline (MAE) are taken from~\cite{saporta2026curia}. The most relevant comparison is \ours{} versus the Baseline (MAE) (same pre-training data and architecture); the FMs are reported for context only.}

\resizebox{\ifdim\width>0.8\linewidth 0.8\linewidth\else \width\fi}{!}{%
\begin{tabular}{l|c|cccccccccccc}
\toprule
\textbf{Model} & \textbf{Avg}. & \shortstack{A1 \\ \scriptsize acc} & \shortstack{A2 \\ \scriptsize acc} & \shortstack{A3 \\ \scriptsize r2} & \shortstack{O1 \\ \scriptsize AUC} & \shortstack{O2 \\ \scriptsize AUC} & \shortstack{O4 \\ \scriptsize cindex} & \shortstack{M4 \\ \scriptsize AUC} & \shortstack{E1 \\ \scriptsize AUC} & \shortstack{E2 \\ \scriptsize AUC} & \shortstack{E3 \\ \scriptsize AUC} & \shortstack{N \\ \scriptsize bacc} & \shortstack{F \\ \scriptsize acc} \\
\midrule
MedImageInsight~\cite{codella2024MedImageInsight} & 83.6 & 86.3 & 83.0 & 80.4 & 79.8 & \textbf{89.8} & 59.3 & 71.2 & 91.5 & 94.4 & 94.6 & \underline{94.4} & 79.0 \\
BioMedCLIP~\cite{zhang2025BioMedCLIP} & 78.3 & 77.8 & 71.9 & 85.0 & 64.1 & 81.3 & 62.3 & 72.8 & 89.2 & 90.1 & 79.3 & 91.7 & 74.1 \\
MedGemma~\cite{sellergren2025medgemma} & 81.8 & 76.2 & 79.5 & 77.7 & 77.9 & \underline{88.7} & \underline{75.3} & 70.1 & 87.9 & 94.0 & 89.7 & 91.2 & 72.7 \\
Curia B~\cite{dancette2025curia} & 86.0 & 88.2 & 92.7 & 87.8 & 82.6 & 77.8 & 74.3 & 78.2 & \textbf{95.1} & 93.5 & 93.8 & 89.0 & 78.8 \\
Curia L~\cite{dancette2025curia} & 84.5 & 89.4 & \underline{95.7} & 86.7 & 84.0 & 77.6 & 60.6 & 74.4 & 94.1 & 91.9 & 87.9 & 92.4 & 79.2 \\
Curia-2 B~\cite{saporta2026curia} & 87.7 & 90.9 & 95.0 & 87.4 & 84.5 & 78.5 & 72.5 & \underline{81.1} & 93.7 & \underline{94.7} & \textbf{99.2} & \textbf{94.7} & 80.2 \\
Curia-2 L~\cite{saporta2026curia} & \textbf{88.6} & \underline{93.2} & \underline{95.7} & \underline{88.9} & \underline{87.1} & 79.3 & \textbf{75.5} & \textbf{84.9} & 94.2 & 93.2 & \underline{98.6} & 92.6 & 79.7 \\
Curia-2 g~\cite{saporta2026curia} & \underline{87.8} & \textbf{94.2} & \textbf{96.5} & \textbf{89.7} & \textbf{87.9} & 75.3 & 74.6 & 79.4 & \underline{94.6} & 93.8 & 96.1 & 92.8 & 79.3 \\
\midrule
CT-CLIP~\cite{ctclip2024} & 51.5 & 38.1 & 20.9 & 22.0 & 63.8 & 56.8 & 58.1 & 43.5 & 69.3 & 42.6 & 69.7 & 69.9 & 63.4 \\
Merlin~\cite{blankemeier2024merlin} & 65.4 & 45.4 & 38.5 & 53.3 & 55.7 & 64.9 & 66.3 & 50.4 & 85.1 & 92.5 & 75.6 & 76.0 & \underline{81.2} \\
Pillar-0~\cite{agrawal2025pillar} & 71.5 & 50.7 & 40.0 & 53.5 & 70.2 & 84.9 & 68.1 & 69.2 & 92.4 & \textbf{98.2} & 56.9 & 91.7 & \textbf{82.4} \\
\midrule
Baseline (MAE) & 73.9 & 70.4 & 74.8 & 77.8 & 81.4 &	75.9 &	49.8 & 71.3 & 86.6 &	86.9 &	65.1 & 84.4 & 61.6 \\
\ours & 74.7 & 65.2 &	68.9 &	79.8 & 80.4 &	78.2 &	63.9 & 71.1 & 86.0 &	83.8 &	71.0 & 86.4 & 62.1 \\
\bottomrule
\end{tabular}
}%
    \label{tab:3dresults}
\end{table}

Beyond dense segmentation, we examine whether the frozen representation transfers to image-level classification and regression tasks, a standard probe of representation quality in the self-supervised literature. 

We evaluate on CuriaBench3D~\cite{saporta2026curia}, a benchmark for volumetric radiology foundation models spanning a range of clinical classification, regression, and survival tasks across CT and MRI; we refer to~\cite{saporta2026curia} for the full task descriptions. Note that CuriaBench3D is composed of public datasets that do not overlap with the pre-training dataset used in our experiments. For each task we attach a lightweight head to the frozen encoder, keeping the backbone fixed as in our segmentation evaluation. Results are shown in Table~\ref{tab:3dresults}.

The foundation models reported on CuriaBench3D (e.g. Curia~\cite{dancette2025curia}, Curia-2~\cite{saporta2026curia}, MedImageInsight~\cite{codella2024MedImageInsight}, or Pillar-0~\cite{agrawal2025pillar}) are explicitly optimized for global, image-level representations (in 2D or 3D depending on the model), whereas our pre-training targets dense features for segmentation. We therefore do not expect to match these models, and report these results only to characterize how our dense-feature pre-training transfers to image-level tasks.

Within this setting, the relevant comparison is between \ours{} and the Baseline (MAE) under the same pre-training data and architecture. \ours{} improves the average score over the baseline ($73.9 \rightarrow 74.7$), with the largest gains on the kidney cancer survival (O4, $49.8 \rightarrow 63.9$) and myocardial infarction (E3, $65.1 \rightarrow 71.0$) tasks, while remaining comparable on the rest. This indicates that the components introduced for dense transfer also yield a modest improvement in the global representation, without having been designed for it. Notably, on a subset of tasks our frozen dense features come close to the dedicated foundation models. For instance on kidney lesion malignancy (O1) ($80.4$ AUC, against $82$--$88$ for the Curia family) and ACL Tear (M4) ($71.1$ AUC, comparable to MedImageInsight at $71.2$), suggesting that segmentation-oriented pre-training already captures clinically relevant image-level signal on these tasks.

\section{Conclusion}
We introduced {\ours}, a new SSL method primarily designed for dense downstream tasks. We developed it across a large number of anatomical sites and two modalities (CT and MRI), demonstrating its broad applicability. More importantly, it opens the possibility of using frozen encoders for 3D medical segmentation, facilitating the integration of such models into clinical practice.

Although frozen-encoder performance does not yet match the full fine-tuning performance of nnU-Net~\cite{nnunet}, which remains the overall state of the art for 3D medical image segmentation, our work begins to close the gap between these two approaches. In a similar vein, and contrary to the findings reported in \cite{revisitingmae}, our experiments suggest that a MAE pre-trained SSL encoder does not consistently outperform the nnU-Net baseline, even under full fine-tuning. We attribute this to the substantially greater difficulty of our task, which involves multi-modality and a large number of anatomical sites, as opposed to the head-and-neck MRI setting considered in \cite{revisitingmae}. It nevertheless remains plausible that such an improvement could emerge with additional data and extended training time.

Furthermore, the variance observed within medical datasets—as illustrated by the results on the MSD-Lung \cite{msd} dataset—makes it difficult to draw firm conclusions regarding dense SSL pre-training. This calls for a rigorous approach involving, in particular, a large number of datasets in order to derive reliable conclusions from the different experiments. This variance in segmentation results across medical datasets may stem from several factors, including the poor quality of certain segmentations, which renders model learning inconsistent; the limited number of training, validation, and test data points; and the inherent difficulty of the task, particularly for lesion-oriented datasets, owing to the variability of the lesions to be segmented.

Several directions remain open for future investigation. Although we developed and evaluated this method using a ResEnc architecture—presented as the state-of-the-art architecture for 3D medical segmentation—the approach remains transferable to other architectures, such as Transformer-based models, on which we believe it could perform equally well, subject to further investigation. In parallel, our preliminary experiments indicate that incorporating a segmentation loss during pre-training tend to improve performance on downstream segmentation tasks; a promising direction would therefore be to explore semi-supervised pre-training, which could further enhance frozen-encoder performance.

\section*{Acknowledgements}
We acknowledge the EuroHPC Joint Undertaking for awarding this project access to the EuroHPC supercomputer LEONARDO, hosted by CINECA (Italy) and the LEONARDO consortium through an EuroHPC AI Factory Access call.
\bibliographystyle{splncs04}
\bibliography{main}
\end{document}